\documentclass[runningheads]{llncs}
\usepackage{graphicx}
\usepackage{amsmath,amssymb} 
\usepackage{color}
\usepackage[width=122mm,left=12mm,paperwidth=146mm,height=193mm,top=12mm,paperheight=217mm]{geometry}

\usepackage{times}
\usepackage{epsfig}
\usepackage[linewidth=1pt]{mdframed}
\usepackage{multirow}
\usepackage[pagebackref=true,breaklinks=true,letterpaper=true,colorlinks,bookmarks=false]{hyperref}

\usepackage{footnote}
\makesavenoteenv{tabular}
\makesavenoteenv{table}

\begin{document}
\pagestyle{headings}
\mainmatter
\def\ECCV18SubNumber{2152}  

\title{A Zero-Shot Framework for Sketch based Image Retrieval} 

\titlerunning{ECCV-18 submission ID \ECCV18SubNumber}


\authorrunning{Sasi Kiran Yelamarthi et al}

\author{Sasi Kiran Yelamarthi$^1$ \and
Shiva Krishna Reddy$^1$ \and
Ashish Mishra \and
Anurag Mittal }


\institute{Indian Institute of Technology Madras, India \\
\email{\{sasikiran1996, shivakrishnam912\}@gmail.com, \{mishra, amittal\}@cse.iitm.ac.in}}

\maketitle

\begin{abstract}
 Sketch-based image retrieval (SBIR) is the task of retrieving images from a natural image database that correspond to a given hand-drawn sketch. Ideally, an SBIR model should learn to associate components in the sketch (say, feet, tail, etc.) with the corresponding components in the image having similar shape characteristics. However, current evaluation methods simply focus only on coarse-grained evaluation where the focus is on retrieving images which belong to the same class as the sketch but not necessarily having the same shape characteristics as in the sketch. As a result, existing methods simply learn to associate sketches with classes seen during training and hence fail to generalize to unseen classes. In this paper, we propose a new benchmark for zero-shot SBIR where the model is evaluated on novel classes that are not seen during training. We show through extensive experiments that existing models for SBIR that are trained in a discriminative setting learn only class specific mappings and fail to generalize to the proposed zero-shot setting. To circumvent this, we propose a generative approach for the SBIR task by proposing deep conditional generative models that take the sketch as an input and fill the missing information stochastically. Experiments on this new benchmark created from the "Sketchy" dataset, which is a large-scale database of sketch-photo pairs demonstrate that the performance of these generative models is significantly better than several state-of-the-art approaches in the proposed zero-shot framework of the coarse-grained SBIR task. 
\end{abstract}

\footnote{Equal Contribution}

\section{Introduction}
The rise in the number of internet users coupled with increased storage capacity, better internet connectivity and higher bandwidths has resulted in an exponential growth in multimedia content on the Web. In particular, image content has become ubiquitous and plays an important role in engaging users on social media as well as customers on various e-commerce sites. With this growth in image content, the information needs and search patterns of users have also evolved. Specifically, it is now common for users to search for images (instead of documents) either by providing a textual description of the image or by providing another image which is similar to the desired image (for example, retrieve all shirts which look similar to the shirt in the query image). The former is known as text based image retrieval and the latter as content based image retrieval \cite{content}. 

The motivation for content based image retrieval can be easily understood by taking an example from online fashion. Here, it is often hard to provide a \textit{textual description} of the desired product but easier to provide a \textit{visual description} in the form of a matching image. The visual description/query need not necessarily be an image but can also be a sketch of the desired product, if no image is available. The user can simply draw the sketch on-the-fly on touch based devices. This convenience in expressing a visual query has led to the emergence of Sketch-based image retrieval (SBIR) as an active area of research \cite{conf/iccv/SBIR1,conf/cvpr/SBIR2,conf/mm/SBIR3,journals/tvcg/SBIR5,journals/cviu/SBIR6,conf/icip/SBIR7,conf/mm/SBIR8,conf/eccv/SBIR9,conf/bmvc/SBIR10,conf/wacv/SBIR11,conf/mir/SBIR12,journals/corr/SBIR13,conf/icip/SBIR14,conf/cvpr/SBIR15,conf/icip/SBIR16}. The primary challenge here is the domain gap between images and sketches wherein sketches contain only an outline of the object and hence have less information compared to images. The second challenge is the large intra-class variance present in sketches due to the fact that humans tend to draw sketches with varied levels of abstraction. 

\begin{figure}
\begin{centering}
\includegraphics[width=7.3cm,height=5cm]{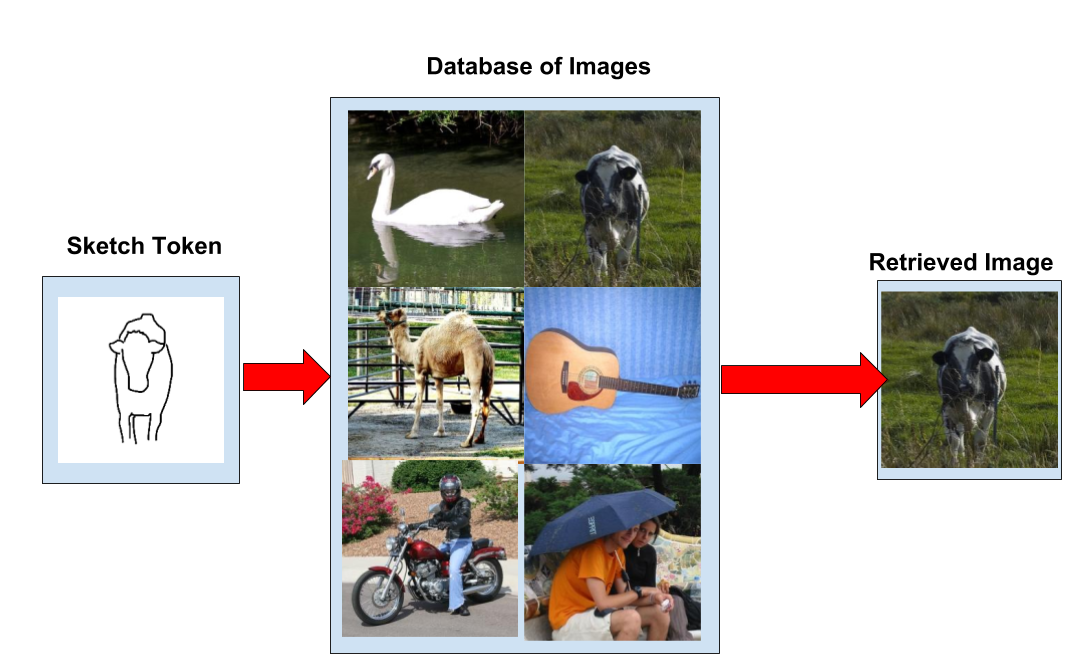}
\par\end{centering}

\caption{Illustration of Sketch based Image Retrieval \label{introFig}}

\end{figure}
 
Ideally, for better generalization, a model for SBIR must learn to discover the alignments between the components of the sketch and the corresponding image. For example, in Figure-\ref{introFig}, we would want the model to associate the head of the cow in the sketch to that in the image. However, current evaluation methodology \cite{journals/corr/LiuSSLS17,Siamese2,journals/tog/Sketchy} that focuses only on class-based retrieval rather than shape or attribute-based retrieval does not explicitly evaluate whether such associations are being learned by the model. Specifically, during evaluation, the model is given credit if it simply fetches an image which belongs to the same class as the sketch. The object in the image need not have the same outline, etc as in the sketch. For example, for the query (sketch) shown in Figure-\ref{introFig}, there is no guarantee that the model fetches the image of the cow with the same number of feet visible or the tail visible, even if it has high evaluation score. 

Thus, a model could possibly achieve good performance by simply learning a class specific mapping from sketches to class labels and retrieve all the images from the same class as that of the query sketch. This is especially so, when the unseen sketches seen at test time belong to the same set of classes as seen during training. Furthermore, existing methods evaluate their models on a set of randomly selected sketches that are withheld during training. However, the images corresponding to the withheld sketches could still occur in the training set, and that would make the task easier. 


One way to discourage such class specific learning is to employ a fine-grained evaluation. For a given sketch, the retrieved results are evaluated by comparing the estimated ranking of images in the database with a human annotated rank list. However, creating such annotations for large datasets such as "Sketchy" \cite{journals/tog/Sketchy} requires extensive human labor. Also, such evaluation metrics are subject to human biases. Indeed, this approach was employed in several prior works such as \cite{conf/cvpr/SBIR15,conf/eccv/SBIR9} In this work, we propose coarse-grained evaluation in the zero-shot setting as a surrogate to fine-grained evaluation to circumvent both these drawbacks. The idea is to test the retrieval on sketches of unseen classes to discourage class-specific learning during training. The evaluation is automatic, \textit{i.e.}, it requires no human labor for each retrieval, apart from having no biases. The model has to learn to associate the latent alignments in the sketch and the image in order to perform well. This is also important from a practical standpoint wherein, in some domains, all possible classes many not be available at training time. For example, new product classes emerge every day in the fashion industry. Thus, the \textit{Zero-Shot Sketch Based Image Retrieval (ZS-SBIR)} task introduced in this paper provides a more realistic setup for the sketch-based retrieval task.

Towards this end, we propose a new benchmark for the ZS-SBIR task by creating a careful split of the Sketchy database (as explained later in the experiments section). We first evaluate several existing SBIR models on this task and observe that the performance of these models drops significantly in the zero-shot setting thereby pointing to class-specific learning occurring in these models. We hypothesize that one reason for this could be that the existing methods are essentially formulated in the discriminative setup, which encourages class specific learning. To circumvent the problems in these existing models, we approach the problem from the point of view of a generative model. Specifically, ZS-SBIR can be considered as the task of generating additional information that is absent in the sketch in order to retrieve similar images. We propose Deep Conditional Generative Models based on Adversarial Autoencoders and Variational Autoencoders for the ZS-SBIR task.  Essentially, our model takes the sketch feature vector as an input and generates a number of possible image vectors by filling the missing information stochastically using generative models. We make use of these generated image feature vectors to retrieve images from the database. Our experiments show that the proposed generative approach performs better than all existing state-of-the-art SBIR models in the zero-shot setting. 
In summary, our main contributions are as follows:


\begin{itemize}
\item We propose the task of Zero-Shot Sketch Based Image Retrieval (ZS-SBIR), which to the best of our knowledge has not been explored before. We argue that this task provides a convenient way to evaluate sketch based retrieval models as a substitute for fine-grained evaluation, while discouraging class specific learning.
\item We provide a benchmark for the proposed problem by creating a careful split of the Sketchy dataset, ensuring that the test classes do not overlap with the 1000 classes of Imagenet \cite{imagenet_cvpr09}. 
\item We propose a novel generative approach to the ZS-SBIR task that achieves better performance in the proposed setting compared to the state-of-the art SBIR models as well as three popular zero-shot image classification algorithms adapted to the ZS-SBIR task. 
\end{itemize}

The paper is organized as follows: In Section-2, we give a brief overview of the state-of-the-art techniques in SBIR and ZSL. Subsequently, in Section-3, we introduce the proposed zero-shot framework and describe the proposed dataset split. Section-4 shows the evaluation of existing state-of-the-art SBIR models in this proposed setting. Section-5 introduces our proposed generative modeling of ZS-SBIR and adaptations of three popular ZSL models to this setting. Finally, in Sections-6, we present an empirical evaluation of these models on the proposed zero shot splits on the Sketchy dataset.   


\section{Related Work}

Since we propose a zero-shot framework for the SBIR task, we briefly review the literature from both sketch-based image retrieval as well as zero-shot learning in this section. 

Conventional pipeline in SBIR involves projecting images and sketches into a common feature space. These features or binary codes extracted from them are used for the retrieval task. Hand-crafted feature based models include the gradient field HOG descriptor proposed by Hu and Collomose \cite{journals/cviu/HOG}, the histogram of edge orientations (HELO) proposed by Saavendra \cite{conf/icip/HELO}, the learned key shapes (LKS) proposed by Saavendra \textit{et.al} \cite{conf/bmvc/LKS} which are used in Bag of Visual Words (BoVW) framework as feature extractors for SBIR. Yu \textit{et.al} \cite{journals/ijcv/CNNinSKETCH} were the first to use Convolutional Neural Networks (CNN) for the sketch classification task. Qi \textit{et.al} \cite{Siamese2} introduced the use of siamese architecture for coarse-grained SBIR. Sangkloy \textit{et.al} \cite{journals/tog/Sketchy} used triplet ranking loss for training the features for coarse-grained SBIR. Yu \textit{et.al} \cite{conf/cvpr/SBIR15} used triplet network for instance level SBIR evaluating the performance on shoe and chair dataset. They use a pseudo fine-grained evaluation where they only look at the position of the correct image for a sketch in the retrieved images. Liu \textit{et.al} \cite{journals/corr/LiuSSLS17} propose a semi-heterogeneous deep architecture for extracting binary codes from sketches and images that can be trained in an end-to-end fashion for coarse-grained SBIR task.

We now review the zero-shot literature. Zero-shot learning in Image Classification \cite{journals/pami/DAP,vinayecml,vinayaaai,mishracvpr} refers to learning to recognize images of novel classes although no examples from these classes are present in the training set. Due to the difficulty in collecting examples of every class in order to train supervised models, zero-shot learning has received significant interest from the research community recently \cite{journals/pami/DAP,conf/nips/CMT,journals/corr/SAE,journals/corr/Latem,conf/icml/ESZSL,conf/cvpr/ZSL1,conf/nips/DeviSE,vinaycvpr}. We refer the reader to \cite{ZSL-GBU} for a comprehensive survey on the subject. Recently, zero shot learning has been gaining increasing attention for a number of other computer vision tasks such as image tagging \cite{conf/sigir/ZSL-ImgTagging,journals/corr/ZSL-ImgTagging}, visual question answering \cite{journals/corr/ZSL-VQA1,journals/corr/RamakrishnanPSM17} action recognition \cite{mishrawacv} etc. To the best of our knowledge, the zero-shot framework has not been previously explored in the SBIR task.


\section{Zero shot setting for SBIR}

We now provide a formal definition of the zero shot setting in SBIR. Let \\ $S=\{(x_i^{sketch},x_i^{img},y_i)|y_i \in \mathcal{Y}\}$ be the triplets of sketch, image and class label where $\mathcal{Y}$ is the set of all class labels in $S$. We partition the class labels in the data into $\mathcal{Y}_{train}$ and $\mathcal{Y}_{test}$  data respectively. Correspondingly, let $S_{tr}=\{(x_i^{sketch},x_i^{img})|y_i \in Y_{train}\}$ and $S_{te}=\{(x_i^{sketch},x_i^{img})|y_i \in Y_{test}\}$ be the partition of S into train and test sets. This way, we partition the paired data into train and test set such that none of the sketches from the test classes occur in the train set. Since the model has no access to class labels, the model needs to learn latent alignments between the sketches and the corresponding images to perform well on the test data. 

Let $D$ be the database of all images and $g_{I}$ be the mapping from images to class labels. We split D into $D_{tr}=\{x_i^{img}\in D|g_I(x_i^{img})\in Y_{train}\}$ and $D_{te}=\{x_i^{img}\in D|g_I(x_i^{img})\in Y_{test}\}$. This is similar to other zero-shot literature \cite{journals/pami/DAP} in image classification. The retrieval model in this framework can only be trained on $S_{tr}$. The database $D_{tr}$ may be used for validating the retrieval results in order to tune the hyper-parameters. Given an $x^{sketch}$ taken from sketches of $S_{te}$, the objective of zero shot setting in SBIR is to retrieve images from $D_{te}$ that belong to same class as that of the query sketch. This evaluation setting ensures that the model can not just learn the mapping from sketches to class labels and retrieve all the images using the label information. The model now has to learn the salient common features between sketches and images and use this to retrieve images for the query that are from the unseen classes.    

\subsection{Benchmark}
Since we are introducing the task of zero-shot sketch based retrieval, there is no existing benchmark for evaluating this setting. Hence, we first propose a new benchmark for evaluation by making a careful split of the "Sketchy" dataset \cite{journals/tog/Sketchy}. Sketchy is a dataset consisting of 75,471 hand-drawn sketches and 12,500 images belonging to 125 classes collected by Sangkloy \textit{et.al} \cite{journals/tog/Sketchy}. Each image has approximately 6 hand-drawn sketches. The original Sketchy dataset uses the same 12,500 images as the database. Liu \textit{et.al} \cite{journals/corr/LiuSSLS17} augment the database with 60,502 images from Imagenet to create a retrieval database with a total of 73,002 images. We use the augmented dataset provided by Liu \textit{et.al} \cite{journals/corr/LiuSSLS17} in this work. 

Next, we partition the 125 classes into 104 train classes and 21 test classes. This peculiar split is not arbitrary. We make sure that the 21 test classes are not present in the 1000 classes of Imagenet \cite{imagenet_cvpr09}. This is done to ensure that researchers can still pre-train their models on the 1000 classes of Imagenet without violating the zero-shot assumption. Such a split was motivated by the recently proposed benchmark for standard datasets used in the zero shot image classification task by Xian \textit{et.al} \cite{ZSL-GBU}. The details of the proposed dataset split are summarized in Table \ref{tab:dataSetStat}. 

\begin{table}
\caption{Statistics of the proposed dataset split of Sketchy database for ZS-SBIR task \label{tab:dataSetStat}}
\begin{center}
\begin{tabular}{|l|c|}
\hline
\textbf{Dataset Statistics} & \textbf{$\#$} \\
\hline\hline
Train classes & 104\\
Test classes & 21\\
Train Images & 10400 \\
Train Sketches & 62787 \\
Avg. sketches per image & 6.03848\\
Test Sketches & 12694 \\
DB images for training & 62549\\
DB images for testing & 10453\\
\hline
\end{tabular}
\end{center}
\end{table}


\section{Limitations of existing SBIR methods}
Next we evaluate whether the existing approaches to the sketch-based image retrival task generalize  well to the proposed zero-shot setting. To this end, we evaluate three state-of-the-art SBIR methods described below on the above proposed benchmark.

\subsection{A Siamese Network}

The Siamese network proposed by Hadsell \textit{et.al} \cite{conf/cvpr/Siamese1} maps both the sketches and images into a common space where the semantic distance is preserved. Let $(S,I,Y=1)$ and $(S,I,Y=0)$ be the pairs of images and sketches that belong to same and different class respectively and $D_{\theta}(S,I)$ be the l2 distance between the image and sketch features where $\theta$ are the parameters of the mapping function. The loss function $L(\theta)$ for training is given by:
\begin{equation} \tag{1} \label{eq:1}
\begin{aligned}
L(\theta)= {} & (Y)\dfrac{1}{2}(D_\theta)^2+(1-Y)\dfrac{1}{2}\{max(0,m-D_{\theta})\}^2
\end{aligned}
\end{equation}
where $m$ is the margin. Chopra \textit{et.al} \cite{Siamese2} and Qi \textit{et.al} \cite{conf/icip/SiameseSketch} use a modified version of the above loss function for training the Siamese network for the tasks of face verification and SBIR respectively, which is given below:
\begin{equation} \tag{2} \label{eq:2}
\begin{aligned}
L(\theta)= {} & (Y)\alpha D_{\theta}^{2}+(1-Y)\beta e^{\gamma D_{\theta}}
\end{aligned}
\end{equation}
where $\alpha = \dfrac{2}{Q}$, $\beta = 2Q$, $\gamma = -\dfrac{2.77}{Q}$ and constant Q is set to the upper bound on $D_\theta$ estimated from the data. We explore both these formulations in the proposed zero-shot setting. We call the former setting as Siamese-1 and the latter as Siamese-2.

\subsection{A Triplet Network}

Triplet loss \cite{conf/cvpr/triplet,journals/tog/Sketchy} is defined in a max-margin framework, where, the objective is to minimize the distance between sketch and positive image that belong to the same class and simultaneously maximize the distance between the sketch and negative image which belong to different classes. The triplet training loss for a given triplet $t(s,p^+,p^-)$ is given by: 
\begin{equation} \tag{3} \label{eq:3}
\begin{aligned}
L_{\theta}(t)= {} & max(0,m+D_{\theta}(s,p^{+})-D_{\theta}(s,p^{-}))
\end{aligned}
\end{equation} where m is the margin and $D_{\theta}$ is the distance measure used.

To sample the negative images during training, we follow two strategies (i) we consider only images from different class and (ii) we consider all the images that do not directly correspond to the sketch, resulting in coarse-grained and fine-grained training of triplet network respectively. We explore both these training methods in the proposed zero-shot setting for SBIR.

\begin{table*}
\caption{Precision and mAP are estimated by retrieving 200 images. - indicates that the authors do not present results on that metric. 1:{Using 128 bit hash codes}
\label{tab:resultsTraditional}}
\begin{center}
\begin{tabular}{|c|c|c|c|c|}
\hline 
\multirow{2}{*}{\textbf{Method}} & \multicolumn{2}{c|}{\textbf{Precision@200}} & \multicolumn{2}{c|}{\textbf{mAP@200}}\tabularnewline
\cline{2-5} 
 & \textbf{Traditional} & \textbf{Zero-Shot} & \textbf{Traditional} & \textbf{Zero-Shot}\tabularnewline
\hline 
\hline
Baseline & - & 0.106 & - & 0.054\tabularnewline
Siamese-1 & - & 0.243 & - & 0.134\tabularnewline
Siamese-2 & 0.690 & 0.251 & 0.518 & 0.149\tabularnewline
Coarse-grained triplet & 0.761 & 0.169 & 0.573 & 0.083\tabularnewline
Fine-grained triplet & - & 0.155 & - & 0.081\tabularnewline
DSH$^1$& 0.866 & 0.153 & 0.783 & 0.059\tabularnewline
\hline 
\end{tabular}
\par\end{center}
\end{table*}

\subsection{Deep Sketch Hashing}

Liu \textit{et.al} \cite{journals/corr/LiuSSLS17} propose an end-to-end framework for learning binary codes of sketches and images which is the current state-of-the-art in SBIR. The objective function consists of the following three terms: (i) cross-view pairwise loss which tries to bring binary codes of images and sketches of the same class to be close (ii) semantic factorization loss which tries to preserve the semantic relationship between classes in the binary codes and (iii) the quantization loss. The overall loss function can be written as 
\begin{equation} \tag{4} \label{eq:4}
\begin{aligned}
L ={} & |W\odot m-B_{I}^{T}B_{S}|^{2}+ \\
      & \lambda\left(|\Phi(Y_{I})-DB_{I}|^{2}+|\Phi(Y_{S})-DB_{S}|^{2}\right)+ \\
      & \gamma\left(|F_{I}(\theta_{I})-B_{I}|^{2}+|F_{S}(\theta_{S})-B_{S}|^{2}\right) \\
      &	\text{s.t. } B_I \in \{-1,1\}^{m\times n_1}, B_S \in \{-1,1\}^{m\times n_2}  
\end{aligned}
\end{equation}
where $B_I$ and $B_S$ are binary codes for images and sketches, $W$ is the cross-view similarity matrix, $\Phi(Y_I)$, $\Phi(Y_S)$ are the word embeddings for image and sketch classes respectively, $D$ is the shared basis for semantic embeddings and $F_I(\theta_I)$ and $F_S(\theta_S)$ are CNNs for image and sketch respectively. 

\subsection{Experiments}
We now present the results of the above described models on our proposed partitions of the "Sketchy" dataset \cite{journals/tog/Sketchy} in order to evaluate them in the zero-shot setting.

While evaluating each model, for a given test sketch, we retrieve the top $K=200$ images from the database that are closest to the sketch in the learned feature space. We use inverse of the cosine similarity as the distance metric. We present the experimental details for the evaluated methods below. 

\noindent \textbf{Baseline: }
We take a VGG-16 network \cite{VGGNet} trained on image classification task on ImageNet-1K \cite{imagenet_cvpr09} as the baseline. The score for a given sketch-image pair is given by the cosine similarity between their VGG features.

\noindent \textbf{Training: } We re-implement the above described models to evaluate them for the ZS-SBIR task. For sanity check, we first reproduce the results on the traditional SBIR task reported in \cite{journals/corr/LiuSSLS17} successfully. We follow the training methodology described in \cite{journals/tog/Sketchy,Siamese2,journals/corr/LiuSSLS17} closely. 

We observe that the validation performance saturates after 20 epochs in case of Siamese network and after 80 epochs for the Triplet network. We also employ data augmentation for training the Triplet network because the available training data is insufficient for proper training. We explore the hyper-parameters via grid search.

In the case of DSH, we use the CNNs proposed by Liu \textit{et.al} \cite{journals/corr/LiuSSLS17} for feature extraction. We train the network for 500 epochs, validating on the train database after every 10 epochs. We explored the hyper-parameters and found that $\lambda = 0.01$ and $\gamma = 10^{-5}$ give the best results similar to the original SBIR training.



The performance of these models on the ZS-SBIR task are shown in Table \ref{tab:resultsTraditional}. For comparative purposes, we also present the performance in the traditional SBIR setting \cite{journals/corr/LiuSSLS17} where the models are trained on the sketch-image pairs of all the classes. We observe that the performance of these models dips significantly, indicating the non-generalizability of existing approaches to SBIR. This performance drop of more than $50\%$ in the zero-shot setting may be due to the fact that these models trained in a discriminative setting may learn to associate the sketches and images to class labels. 

Among the compared methods we notice that the Siamese network preforms the best among the existing SBIR methods in the zero-shot setting. We also observe that the Triplet loss gives poorer performance compared to the Siamese network. This can be attributed to the presence of only about $60,000$ images during training, which is not sufficient for properly training a triplet network as observed by Schroff \textit{et.al} \cite{conf/cvpr/triplet}. We also observe that the coarse-grained training of triplet performs better compared to fine-grained triplet. This may be because the fine-grained training considers all the images other than those that correspond directly to the sketch as negative samples making the training harder. 

Our next observation is that DSH, which is the state-of-the-art model in SBIR does not perform well compared to either Siamese or Triplet networks in ZS-SBIR task. This may be due to the fact that the semantic factorization loss in DSH takes only the training class embeddings into account and does not reduce the semantic gap for the test classes. 

Thus, one can claim that there exists a problem of class-based learning inherent in the existing models, which leads to inferior performance in the ZS-SBIR task.

\section{Generative Models for ZS-SBIR}

Having noticed that the existing approaches do not generalize well to the ZS-SBIR task, we now propose the use of generative models for the ZS-SBIR task. The motivation for such an approach is that  while a sketch gives a basic outline of the image, additional details could possibly be generated from the latent prior vector via a generative model. This is inline with the recent work on similar image translation tasks \cite{pix2pix2016,journals/corr/styleTransfer,text2pic} in computer vision.

Let $G_\theta$ model the probability distribution of the image features ($x_{img}$) conditioned on the sketch features ($x_{sketch}$) and parameterized by $\theta$, i.e  $\mathbb{P}({x_{img}|x_{sketch};\theta})$. $G_\theta$ is trained using paired data of sketch-image pairs from the training classes. Since we do not provide the model with class label information, it is hoped that the model learns to associate the characteristics of the sketch such as the general outline, local shape, etc with that of the image. We would like to emphasize here that $G_\theta$ is trained to generate image features but not the images themselves using the sketch.  We consider two popular generative models: Variational Autoencoders \cite{VAE,conf/nips/CVAE} and Adversarial Autoencoders \cite{AAE} as described below: 
\begin{figure*}
\begin{center}
\begin{tabular}{|c|c|}
\hline 
\includegraphics[width=6cm]{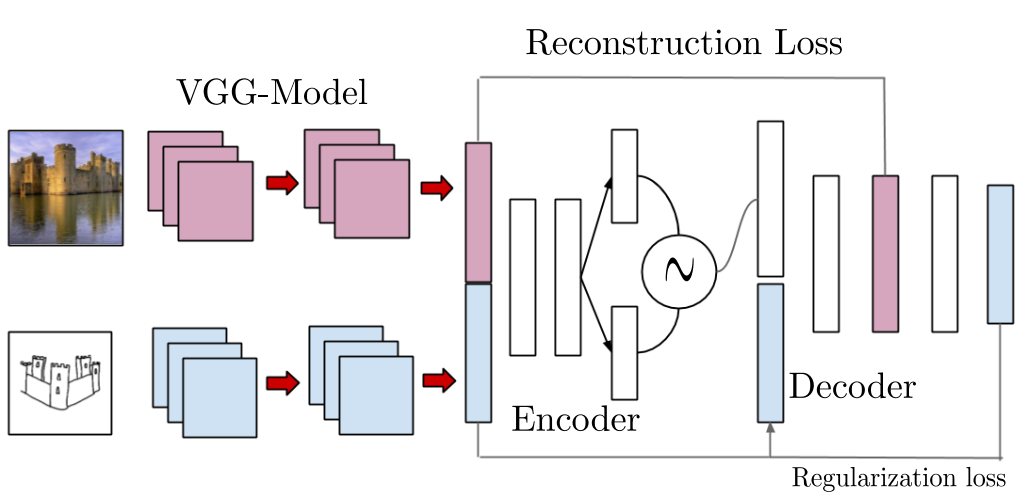} & \includegraphics[width=6cm]{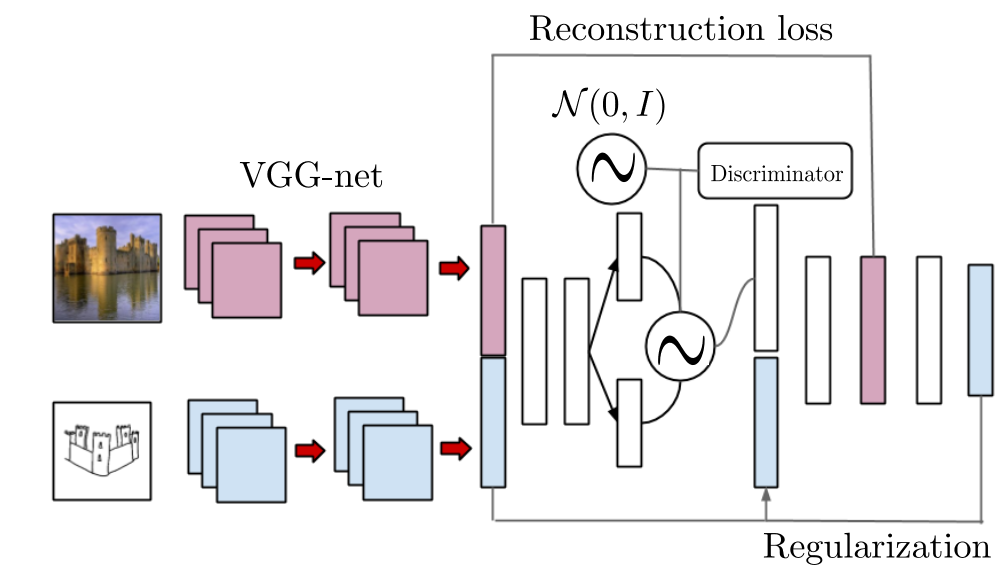}\tabularnewline
\hline 
\end{tabular}
\par\end{center}
\caption{The architectures of CVAE and CAAE are illustrated in the left and right diagrams respectively}
\label{fig:model}
\end{figure*}

\subsection{Variational Autoencoders}
The Variational Autoencoders (VAE) \cite{VAE} map a prior distribution on a hidden latent variable $p(z)$ to the data distribution $p(x)$. The intractable posterior $p(z|x)$ is approximated by the variational distribution $q(z|x)$ which is assumed to be Gaussian in this work. The parameters of the variational distribution are estimated from $x$ via the encoder which is a neural network parameterized by $\phi$. The conditional distribution $p(x|z)$ is modeled by the decoder network parameterized by $\theta$. Following the notation in \cite{VAE}, the variational lower bound for $p(x)$ can be written as: 
\begin{equation} \tag{5} \label{eq:5}
\begin{aligned}
p(x) 	& \geq\mathcal{L}(\phi,\theta;x)\\
 		& =-D_{KL}\left(q_{\phi}(z|x)||p_{\theta}(z)\right)+\mathbb{E}_{q_{\phi}(z|x)}\left[\log p_{\theta}(x|z)\right]
\end{aligned}
\end{equation}
Similarly, it is possible to model the conditional probability $p(x|y)$ as proposed by \cite{conf/nips/CVAE}. In this work, we model the probability distribution over images conditioned on the sketch i.e $P\left(x_{img}|x_{sketch}\right)$. The bound now becomes: 

$\mathcal{L}(\phi,\theta;x_{img},x_{sketch})=$
\begin{align*} \tag{6} \label{eq:6}
-D_{KL}\left(q_{\phi}\left(z|x_{img},x_{sketch}\right)||p_{\theta}\left(z|x_{sketch}\right)\right)+\\
\mathbb{E}\left[\log p_{\theta}\left(x_{img}|z,x_{sketch}\right)\right]
\end{align*}

Furthermore, to encourage the model to preserve the latent alignments of the sketch, we add the reconstruction regularization to the objective. In other words, we force the reconstructibility of the sketch features from the generated image features via a one-layer neural network $f_{NN}$ with parameters $\psi$. All the parameters $\theta,\psi \& \phi$  are trained end-to-end. The regularization loss can be expressed as
\[ \tag{7} \label{eq:7}
\mathcal{L}_{recons} = \lambda.\left|\left|f_{NN}(\widehat{x}_{img})-x_{sketch}\right|\right|_{2}^{2}
\]
Here, $\lambda$ is a hyper-parameter which is to be tuned. The architecture of the conditional variational autoencoder used is shown in Figure \ref{fig:model}. We call this CVAE from here on. 
\subsection{Adversarial Autoencoders}
Adversarial Autoencoders \cite{AAE} are similar to the variational autoencoder, where the KL-Divergence term is replaced with an adversarial training procedure. Let $E, D$ be the encoder and decoder of the autoencoder respectively. E maps input $x_{img}$ to the parameters of the hidden latent vector distribution $P(z|x_{img})$, whereas, D maps the sampled $z$ to $x_{img}$ (both are conditioned on the sketch vector $x_{sketch}$). We have an additional network $\mathcal{D}$: the discriminator. The networks E \& D try to minimize the following loss: 
\[ \tag{8} \label{eq:8}
\mathbb{E}_{z}\left[\log p_{\theta}\left(x_{img}|z,x_{sketch}\right)\right]+\mathbb{\mathbb{E}}_{x_{img}}\left[\log\left(1-\mathcal{D}(E(x_{img}))\right)\right]
\]

\noindent The discriminator $\mathcal{D}$ tries to maximize the following similar to the original GAN formulation \cite{conf/nips/GAN}: 
\[ \tag{9} \label{eq:9}
\mathbb{E}_{z}\left[\log\left[\mathcal{D}(z)\right]\right]+\mathbb{E}_{x_{img}}\left[\log\left[1-\mathcal{D}\left(E(x_{img})\right)\right]\right]
\]
\noindent We add the reconstructibility regularization described in the above section to the loss of the encoder. The architecture of the adversarial autoencoder used is shown in Figure \ref{fig:model}. We call this CAAE from here on.

\subsection{Retrieval Methodology}
\noindent $G_\theta$ is trained on the sketch-image feature pairs from the seen classes. During test time, the decoder part of the network is used to generate a number of image feature vectors $x_{gen}^I$ conditioned on the test sketch by sampling latent vectors from the prior distribution $p(z)=\mathcal{N}(0,I)$. For a test sketch $x_{S}$ corresponding to a test class, we generate the set $\mathcal{I}_{x_{S}}$ consisting of N (a hyper-parameter) such samples of $x_{gen}^I$. We then cluster these generated samples $\mathcal{I}_{x_{S}}$ using K-Means clustering and obtain K cluster centers $C_1,C_2,\dots,C_k$ for each test sketch. We retrieve 200 images $x_{db}^{I}$ from the image database based on the following distance metric:
\[ \tag{10} \label{eq:10}
\mathcal{D}(x_{I}^{db},\mathcal{I}_{x_{S}})=min_{k=1}^{K}cosine\left(\theta(x_{I}^{db}),\theta(C_{k})\right)
\]
\noindent where $\theta$ is the VGG-16 \cite{VGGNet} function. We empirically observe that $K=5$ gives the best results for retrieval. Other distance metrics typically used in clustering were considered but this gave the best results.

\subsection{Experiments}
We conduct an evaluation of the generative models on the proposed zero-shot setting and compare the results with those of existing methods in SBIR. We use the same metrics i.e Precision and mAP, for evaluation. We use the VGG-16 \cite{VGGNet} model pre-trained on the Imagenet-1K dataset to obtain 4096 dimensional features for images. To extract the sketch features, we tune the network for sketch classification task using only the training sketches. We observed that this training gives only a marginal improvement in the performance and is hence optional.  
\subsubsection*{Baselines}
Along with the state-of-the-art models for the SBIR task, we consider three popular algorithms \cite{ZSL-GBU} from the zero-shot image classification literature that do not explicitly use class label information and can be easily adopted to the zero-shot SBIR task. Let $(X_I, X_S)\in (\mathbb{R}^{N\times d_I}, \mathbb{R}^{N\times d_S})$ represent the image and sketch feature pairs from the training data respectively. We learn a mapping $f$ from sketch features to image features, i.e $f:\mathbb{R}^d_{I}\rightarrow \mathbb{R}^d_{S} $ where $d_I, d_S$ are the dimensions of the image and sketch vectors respectively.  We describe these models below:

\noindent \textbf{Direct Regression: }The ZS-SBIR task is formulated as a simple regression problem, where each feature of the image feature vector is learnt from the sketch features. This is similar to the Direct Attribute prediction \cite{journals/pami/DAP} which is a widely used baseline for zero-shot image classification. 

\noindent \textbf{Embarrassingly Simple Zero-Shot Learning: }ESZSL was introduced by Romera-Paredes \& Torr \cite{conf/icml/ESZSL} as a method of learning bilinear compatibility matrix between images and attribute vectors in the context of zero-shot classification. In this work, we adapt the model to the ZS-SBIR task by mapping the sketch features to the image features using parallel training data from the train classes. The objective is to estimate $W\in \mathbb{R}^{d_S\times d_I}$ that minimizes the following loss: 
\[ \tag{11} \label{eq:11}
\left\lvert\left\lvert X_{S}W-X_{I}\right\lvert\right\lvert_{F}^{2}+\gamma\left\lvert \left\lvert X_{I}W^{T}\right\lvert \right\lvert _{F}^{2}+\lambda\left\lvert \left\lvert X_{S}W\right\lvert \right\lvert _{F}^{2}+\beta\left\lvert \left\lvert W\right\lvert \right\lvert _{F}^{2}
\]
 where $\gamma,\,\lambda,\,\beta$ are hyper-parameters. 
 
\noindent \textbf{Semantic Autoencoder: } The Semantic Autoencoder (SAE) \cite{journals/corr/SAE} proposes an autoencoder framework to encourage the re-constructibility of the sketch vector from the generated image vector. The loss term is given by: 
\[ \tag{12} \label{eq:12}
\left\lvert \left\lvert X_{I}-X_{S}W\right\lvert \right\lvert _{F}^{2}+\lambda\left\lvert \left\lvert X_{I}W^{T}-X_{S}\right\lvert \right\lvert _{F}^{2}
\]
We would like to note here that SAE, though simple, is currently the state-of-the-art among published models for zero-shot image classification task to the best of our knowledge.

\subsubsection*{Training}
We use Adam optimizer \cite{journals/corr/Adam} with learning rate $\alpha = 2\times 10^{-4}$, $\beta_1 = 0.5$, $\beta_2 = 0.999$ and a batch size of 64 and 128 for training the CVAE and CAAE respectively. We observe that the validation performance saturates at 25 epochs for the CVAE model and at 6000 iterations for the CAAE model. While training CAAE, we train the discriminator for 32 iterations for each training iteration of the encoder and decoder. We found that $N=200$ i.e generating 200 image features for a given input sketch gives optimal performance and saturates afterwards. The reconstructibility parameter $\lambda$ is set via cross-validation.

SAE has a single hyper-parameter and is solved using the Bartels-Stewart algorithm \cite{bartels1972solution}. ESZSL has three hyper parameters $\gamma, \lambda$ $ \& $ $ \beta$. We set $\beta=\gamma\lambda$ following the authors to get a closed form solution.  We tune these hyper-parameters via a grid search from $10^{-6}$ to $10^7$.

\begin{table}
\caption{The Precision and MAP evaluated on the retrieved 200 images in ZS-SBIR on the proposed split \label{tab:results}}
\begin{center}
\begin{tabular}{|c|c|c|c|}
\hline 
\textbf{Type} & \textbf{Evaluation Methods} & \textbf{\textbf{Precision@200}}  & \textbf{mAP@200}\tabularnewline
\hline 
\hline 
\multirow{6}{*}{SBIR methods} & \multicolumn{1}{c|}{Baseline} & \multicolumn{1}{c|}{0.106} & \multicolumn{1}{c|}{0.054}\tabularnewline
                                & \multicolumn{1}{c|}{Siamese-1} & \multicolumn{1}{c|}{0.243} & \multicolumn{1}{c|}{0.134}\tabularnewline  
                                & \multicolumn{1}{c|}{Siamese-2} & \multicolumn{1}{c|}{0.251} & \multicolumn{1}{c|}{0.149}\tabularnewline 
                                & \multicolumn{1}{c|}{Coarse-grained triplet} & \multicolumn{1}{c|}{0.169} & \multicolumn{1}{c|}{0.083}\tabularnewline
                                & \multicolumn{1}{c|}{Fine-grained triplet} & \multicolumn{1}{c|}{0.155} & \multicolumn{1}{c|}{0.081}\tabularnewline 
                                & \multicolumn{1}{c|}{DSH} & \multicolumn{1}{c|}{0.153} & \multicolumn{1}{c|}{0.059}\tabularnewline 
\hline 
\hline 
\multirow{3}{*}{ZSL methods} & \multicolumn{1}{c|}{Direct Regression} & \multicolumn{1}{c|}{0.066} & \multicolumn{1}{c|}{0.022}\tabularnewline
                                & \multicolumn{1}{c|}{ESZSL} & \multicolumn{1}{c|}{0.187} & \multicolumn{1}{c|}{0.117}\tabularnewline  
                                & \multicolumn{1}{c|}{SAE} & \multicolumn{1}{c|}{0.238} & \multicolumn{1}{c|}{0.136}\tabularnewline

\hline 
\hline 
\multirow{2}{*}{Ours} & \multicolumn{1}{c|}{CAAE} & \multicolumn{1}{c|}{\textbf{0.260}} & \multicolumn{1}{c|}{\textbf{0.156}}\tabularnewline
                                & \multicolumn{1}{c|}{CVAE} & \multicolumn{1}{c|}{\textbf{0.333}} & \multicolumn{1}{c|}{\textbf{0.225}}\tabularnewline
\hline 
\end{tabular}
\par\end{center}
\end{table}

\section{Results}
\fboxsep=0mm
\fboxrule=2pt

\begin{figure}[!ht]
\begin{center}
\begin{mdframed}
\begin{tabular}{ccccccc}
\includegraphics[width=1.50cm, height=1.50cm]{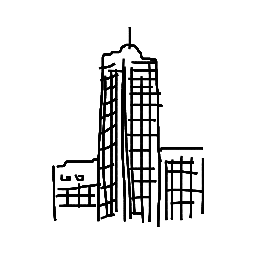} & \includegraphics[width=1.50cm, height=1.50cm]{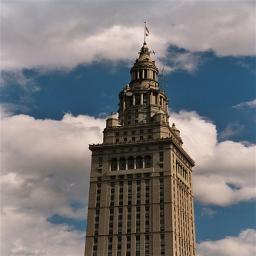} & \includegraphics[width=1.50cm, height=1.50cm]{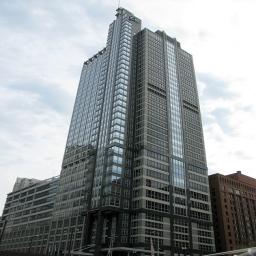} & \includegraphics[width=1.50cm, height=1.50cm]{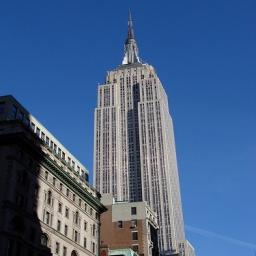} & \includegraphics[width=1.50cm, height=1.50cm]{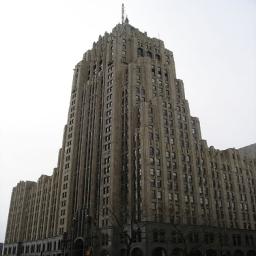} & \includegraphics[width=1.50cm, height=1.50cm]{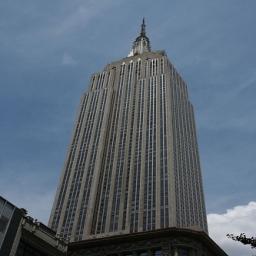} & \includegraphics[width=1.50cm, height=1.50cm]{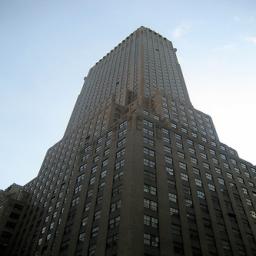}\tabularnewline
\includegraphics[width=1.50cm, height=1.50cm]{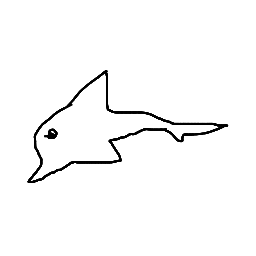} & \includegraphics[width=1.50cm, height=1.50cm]{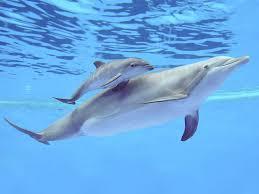} & \includegraphics[width=1.50cm, height=1.50cm]{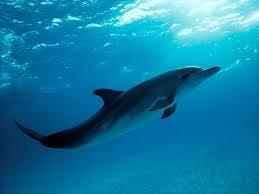} & \includegraphics[width=1.50cm, height=1.50cm]{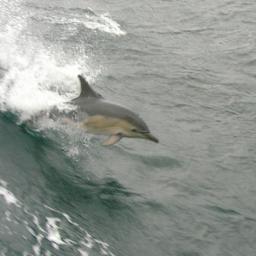} & \includegraphics[width=1.50cm, height=1.50cm]{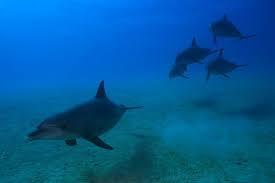} & \includegraphics[width=1.50cm, height=1.50cm]{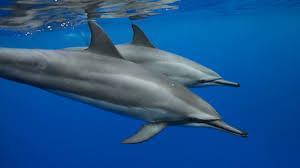} & \includegraphics[width=1.50cm, height=1.50cm]{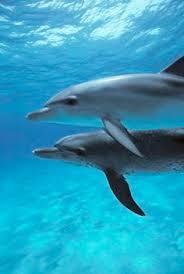}\tabularnewline
\includegraphics[width=1.50cm, height=1.50cm]{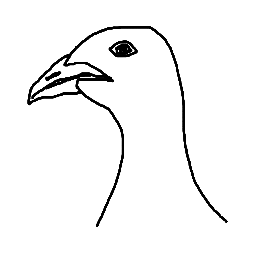} & \includegraphics[width=1.50cm, height=1.50cm]{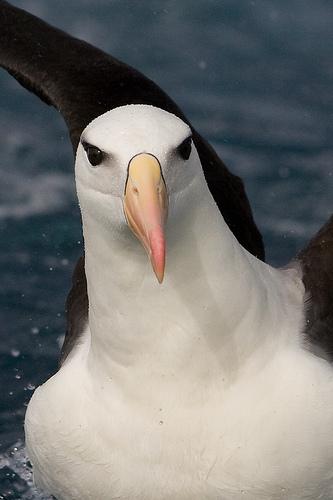} & \includegraphics[width=1.50cm, height=1.50cm]{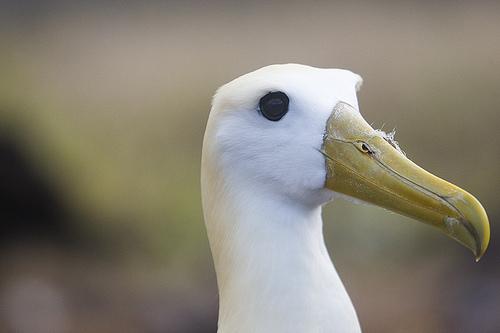} & \includegraphics[width=1.50cm, height=1.50cm]{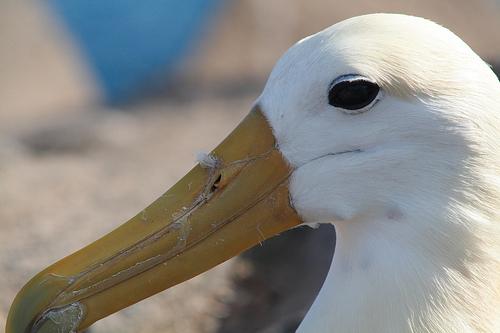} & \includegraphics[width=1.50cm, height=1.50cm]{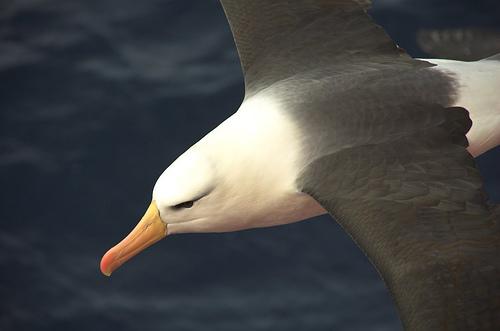} & \includegraphics[width=1.50cm, height=1.50cm]{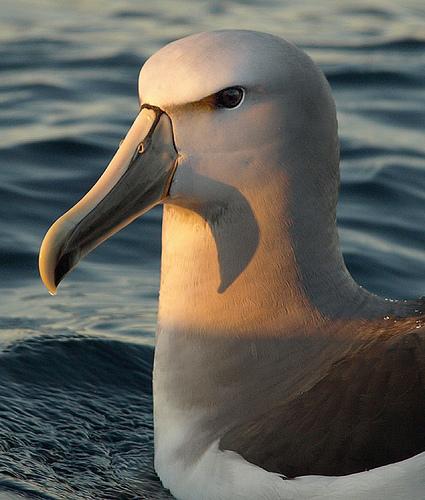} & \includegraphics[width=1.50cm, height=1.50cm]{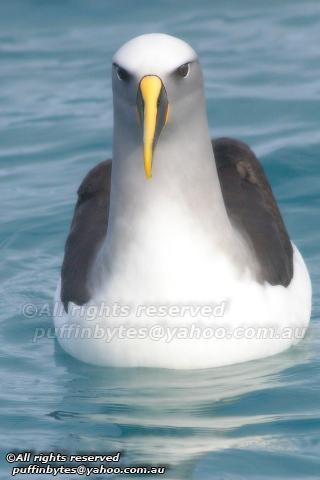}\tabularnewline
\includegraphics[width=1.50cm, height=1.50cm]{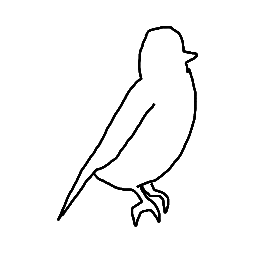} & \includegraphics[width=1.50cm, height=1.50cm]{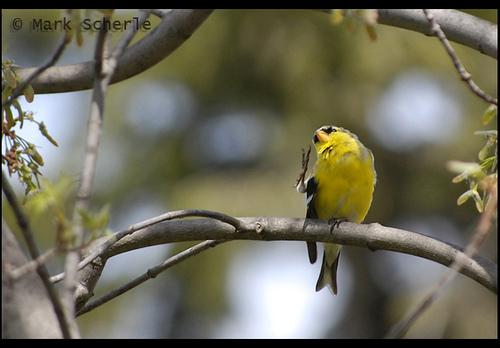} & \fcolorbox{red}{yellow}{\includegraphics[width=1.50cm, height=1.50cm]{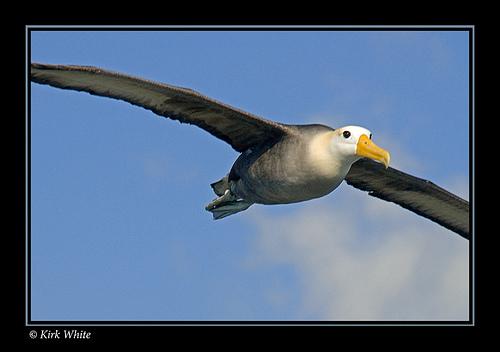}} & \includegraphics[width=1.50cm, height=1.50cm]{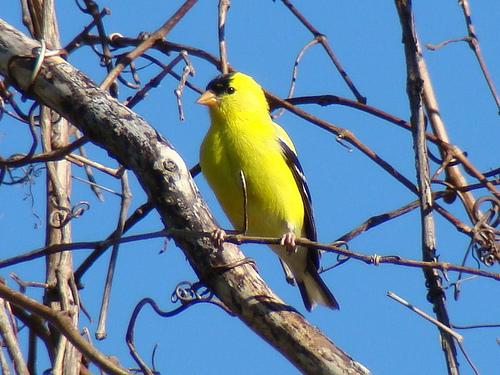} & \includegraphics[width=1.50cm, height=1.50cm]{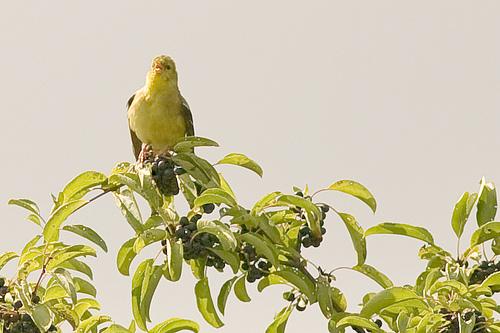} & \includegraphics[width=1.50cm, height=1.50cm]{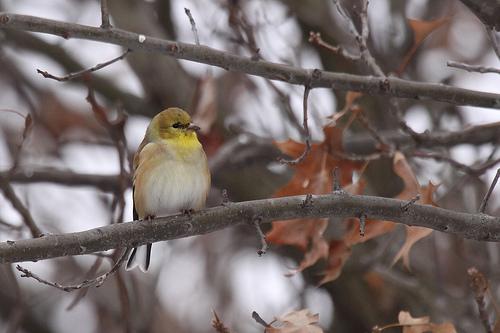} & \includegraphics[width=1.50cm, height=1.50cm]{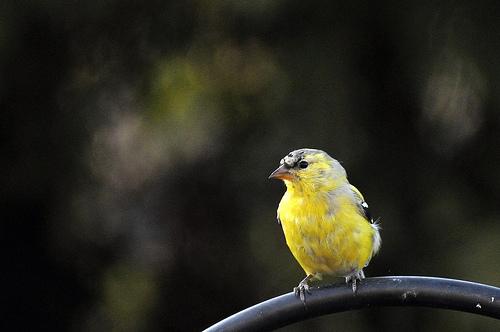}\tabularnewline
\includegraphics[width=1.50cm, height=1.50cm]{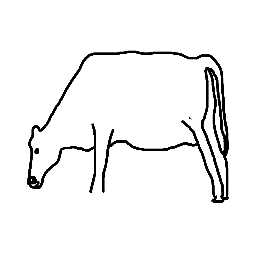} & \includegraphics[width=1.50cm, height=1.50cm]{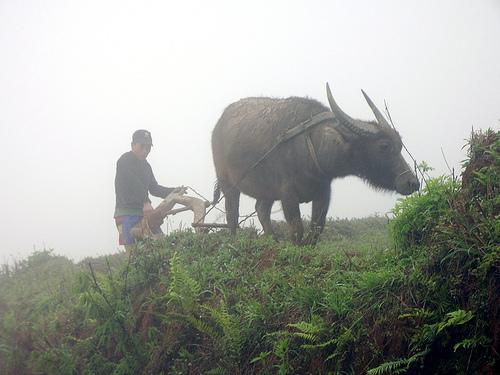} & \includegraphics[width=1.50cm, height=1.50cm]{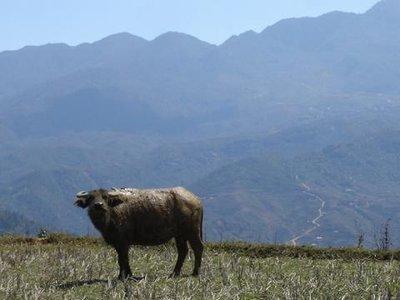} & \includegraphics[width=1.50cm, height=1.50cm]{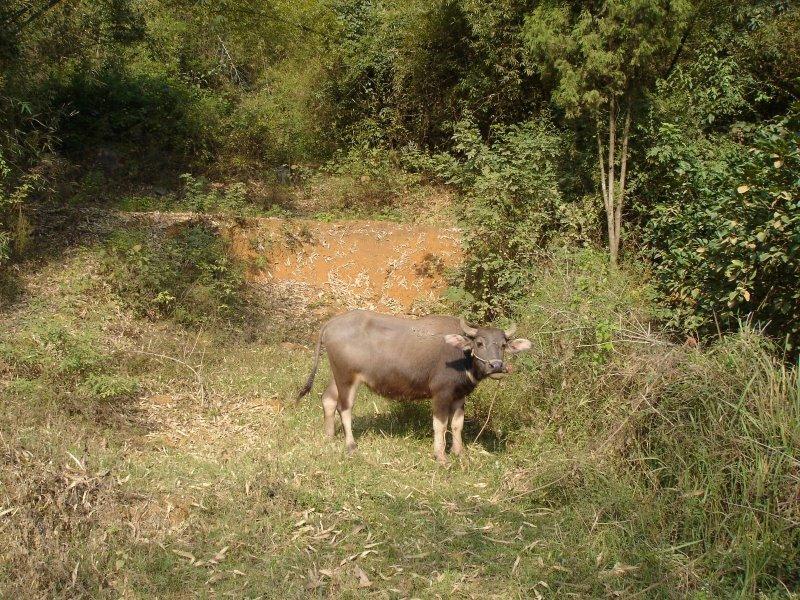} & \includegraphics[width=1.50cm, height=1.50cm]{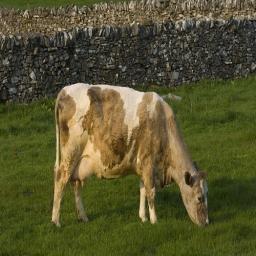} & \includegraphics[width=1.50cm, height=1.50cm]{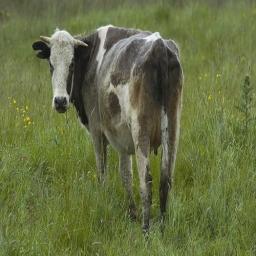} & \fcolorbox{red}{yellow}{\includegraphics[width=1.50cm, height=1.50cm]{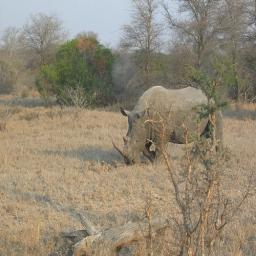}}\tabularnewline

\end{tabular}

\end{mdframed}
\end{center}
\caption{Top 6 images retrieved for some input sketches using CVAE in the proposed zero-shot setting. Note that these sketch classes have never been encountered by the model during training. The red border indicates that the retrieved image does not belong to sketch's class. However, we would like to emphasize that the retrieved false positives do match the outline of the sketch}
\label{fig:examples}
\end{figure}

The results of the evaluated methods for ZS-SBIR are summarized in Table \ref{tab:results}. As observed in section-4.4, existing SBIR models perform poorly in the ZS-SBIR task. Both the proposed generative models out-perform the existing models indicating better latent alignment learning in the generative approach.

\noindent \textbf{Qualitative Analysis: } We show some of the retrieved images for sketch inputs of the unseen classes using the CVAE model in ZS-SBIR in Figure \ref{fig:examples}. We observe that the retrieved images closely match the outline of the sketch. We also observe that our model makes visually reasonable mistakes in the case of false positives wherein the retrieved images do have a significant similarity with the sketch even though they belong to a different class. For instance, in the last example the false positive that belongs to the class rhinoceros has a similar outline as that of the sketch. These may be considered not as an error but rather as a positive retrieval, but can only be evaluated qualitatively by an arduous manual task and may be attributed to data bias. 

\begin{figure}
\begin{center}
\begin{tabular}{|cc|}
\hline 
\includegraphics[width=4cm,height=4cm]{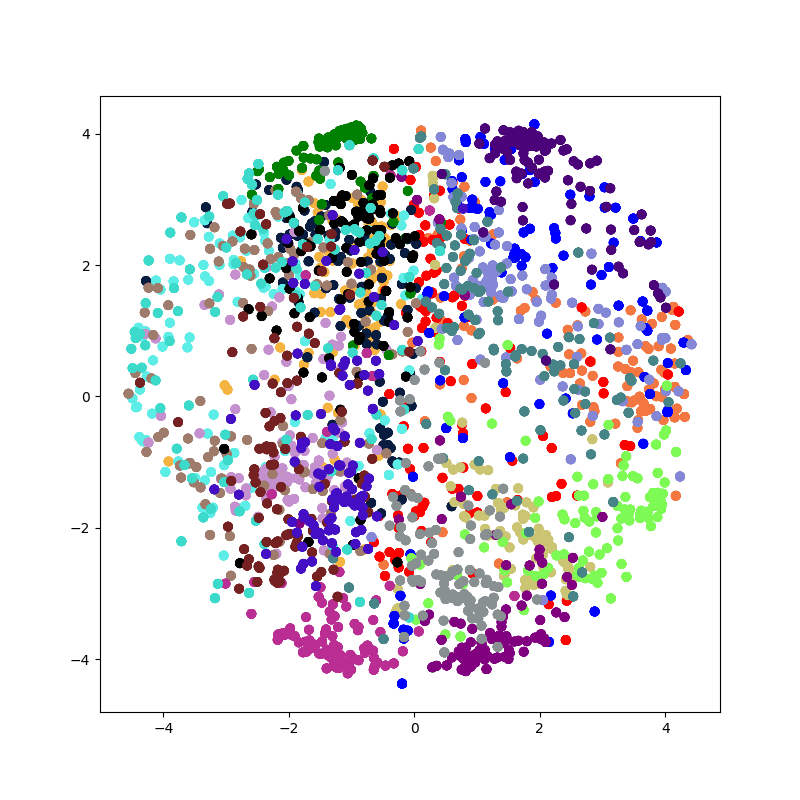} & \includegraphics[width=4cm,height=4cm]{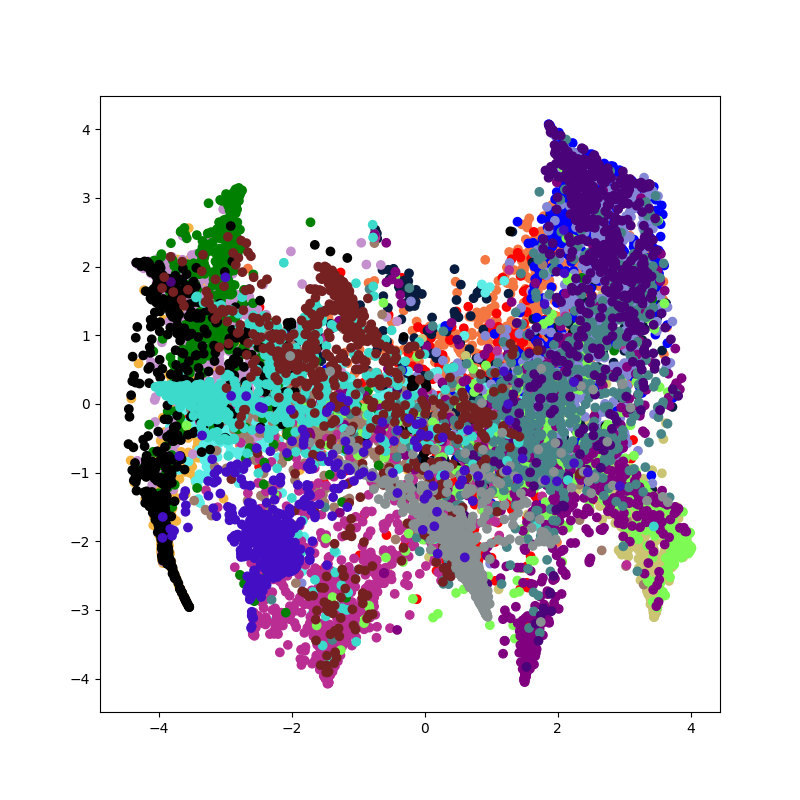}\tabularnewline
\hline 
\end{tabular}
\par\end{center}
\caption{T-SNE visualization of generated image features. Test data features are presented on the left and the predicted image features are on the right. Each color represents a particular class  \label{tSNE}}
\end{figure}

\noindent \textbf{Feature visualization: } To understand the kinds of features generated by the model, we visualize the generated image features of the test sketches in Figure \ref{tSNE} via the t-sne method. We make two observations, (i) the generated features are largely close to the true test image features (ii) multiple modalities of the distribution are captured by our model.

\noindent \textbf{Performance Comparisons: }Comparison among the current state-of-the-art models in the zero-shot setting of SBIR was already done in Section 4.4.  

Direct regression from sketch to image feature space gives a precision value of 0.066. This serves as a baseline to evaluate other explicitly imposed regularizations in ESZSL and SAE. Our first observation is that the simple zero-shot learning models adapted to the ZS-SBIR task perform better than two state-of-the-art sketch based image retrieval models i.e Triplet network and DSH. SAE, which is the current state-of-the-art for zero-shot image classification, achieves the best performance among all the prior methods considered. SAE maps the sketches to images and hence generates a single image for a given sketch. This is similar to our proposed models except that our models generate a number of samples for a single sketch by filling the missing details from the latent distribution. Furthermore our model is non-linear whereas SAE is a simple linear projection. We believe that these generalizations over the SAE in our model leads to superior performance

Among the two models proposed, we observe that the CVAE models performs significantly better than the CAAE model. This may be attributed to the issue of instability while training adversarial models. We observe that the training error of the CVAE models is much more smoother compared to the CAAE model. We observe that using the reconstruction loss leads to a $3\%$ improvement on the precision. We further apply these proposed generative models to the zero-shot image classification task and achieve significant improvements over the state-of-the-art methods. This work has been submitted to a parallel conference and has been included in the Supplementary Material \cite{mishracvpr}.

\section{Conclusion}
We identified major drawbacks in current evaluation schemes in sketch-based image retrieval (SBIR) task. While coarse-grained evaluation suffers from class-specific learning, fine-grained evaluation requires arduous manual labor while also suffering from human biases. To this end, we pose the problem of sketch-based retrieval in a zero-shot framework (ZS-SBIR). By making a careful split in the "Sketchy" dataset, we provide a benchmark for this task. We then evaluate current state-of-the-art SBIR models in this framework and show that the performance of these models drop significantly, thus exposing the class-specific learning which is inherent to these discriminative models. 

We also extend the existing zero-shot image classification methods to this ZS-SBIR task. We then pose the SBIR problem as a generative task and propose two conditional generative models which achieve significant improvement over the existing methods. 


\clearpage

\bibliographystyle{splncs}
\bibliography{egbib}
\end{document}